\documentclass{article}

\newif\ifarxiv
\arxivtrue   

\PassOptionsToPackage{numbers, compress}{natbib}

\ifarxiv
  \usepackage[preprint]{neurips_2026}
\else
  \usepackage{neurips_2026}
\fi

\usepackage[utf8]{inputenc}
\usepackage[T1]{fontenc}
\usepackage{hyperref}
\usepackage{url}
\usepackage{booktabs}
\usepackage{amsfonts}
\usepackage{amsmath}
\usepackage{nicefrac}
\usepackage{microtype}
\usepackage{xcolor}
\usepackage{graphicx}
\usepackage{tabularx}
\usepackage{makecell}
\usepackage{tikz}
\usetikzlibrary{arrows.meta}
\usepackage{algorithmic}
\usepackage{algorithm}
\usepackage{caption}

\newcolumntype{Y}{>{\centering\arraybackslash}X}

\definecolor{pBlue}{HTML}{4472C4}
\definecolor{pOrange}{HTML}{ED7D31}
\definecolor{pPurple}{HTML}{7030A0}
\definecolor{pRed}{HTML}{C00000}

\title{Sparse Autoencoders enable Robust and Interpretable Fine-tuning of CLIP models}

\author{
    Fabian Morelli \\
    University of Tübingen
    \thanks{\texttt{fabian.morelli@student.uni-tuebingen.de}}
\And
    Arnas Uselis\\
    University of Tübingen
\And
    Ankit Sonthalia\\
    University of Tübingen
\And
    Seong Joon Oh\\
    University of Tübingen\\
    KAIST
}

\begin{document}
\maketitle

\begin{abstract}
  Large-scale pre-trained vision-language models like CLIP demonstrate remarkable zero-shot performance across diverse tasks. However, fine-tuning these models to improve downstream performance often degrades robustness against distribution shifts. Recent approaches have attempted to mitigate this trade-off, but often rely on computationally expensive text-guidance. We propose a novel method for robust fine-tuning, SAE-FT, which operates only on the model's visual representations. SAE-FT regularizes changes to these representations by penalizing the addition and removal of semantically meaningful features identified by a Sparse Autoencoder trained on the pre-trained model. This constraint prevents catastrophic forgetting and makes the fine-tuning process interpretable, enabling direct analysis of semantic changes. SAE-FT is both mechanistically transparent and computationally efficient, matching or exceeding state-of-the-art performance on ImageNet and its associated distribution shift benchmarks. Code is publicly available at: https://github.com/Fabian-Mor/sae-ft
\end{abstract}

\section{Introduction}\label{sec: intro}
Contrastive Language-Image Pre-training (CLIP) \cite{radfordLearningTransferableVisual2021} enables the training of large-scale vision-language models on diverse image-caption datasets. These models can subsequently be used for the zero-shot classification of images and generalize to a wide range of tasks, without task-specific training. When evaluated on distribution shifts, CLIP models are more robust than models trained directly on the individual datasets \cite{shi2023effective}.

The performance of the zero-shot model can be further improved by fine-tuning on downstream datasets. While fine-tuning of CLIP models does improve in-distribution (ID) performance, the out-of-distribution (OOD) performance measured with distribution shifts often decreases \cite{wortsmanRobustFinetuningZeroshot2022, kumar2022fine}. \ifarxiv This undesired property has led to increased efforts to understand the fine-tuning process and prevent this degradation in OOD performance. One of the first methods for such robust fine-tuning is WiSE-FT \cite{wortsmanRobustFinetuningZeroshot2022}, which averages the weights of the fine-tuned model with the zero-shot model. While WiSE-FT simplifies the process by effectively ignoring the text encoder, more recent approaches try to improve results by actively fine-tuning both the vision and text components \cite{goyal2022finetunelikepretrainimproved}. However, these methods often also rely on complex data manipulations to succeed. For instance, they may require retrieving additional context information \cite{mao2022contextawarerobustfinetuning} or injecting synthetic features into the text prompts \cite{kim2025starftrobustfinetuningzeroshot}. This dependence introduces external priors and data engineering that complicate the fine-tuning pipeline.

\begin{figure}
    \centering
    \includegraphics[width=0.9\linewidth]{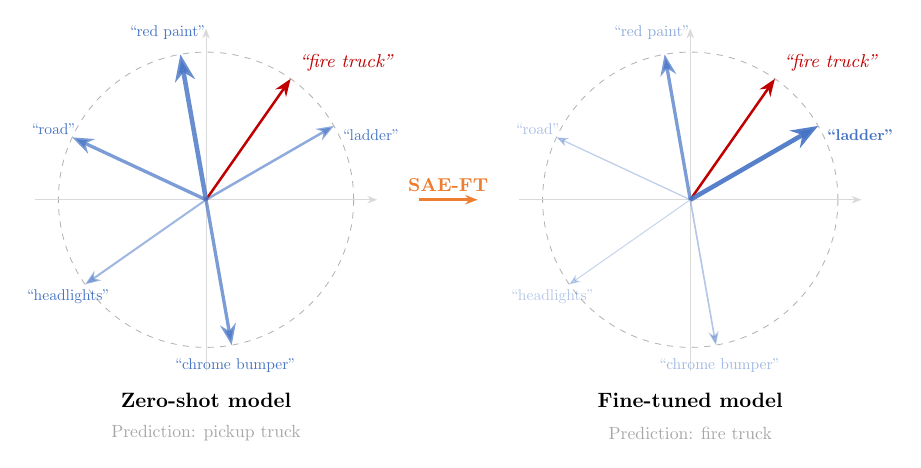}
    \caption{Intuition behind SAE-FT. A Sparse Autoencoder trained on the zero-shot model decomposes CLIP's vision representations into semantically meaningful feature directions (\textcolor{pBlue}{blue}), while the target class embedding (\textcolor{pRed}{red}) defines the direction used for classification. Line thickness indicates each feature's activation strength. The zero-shot model (left) spreads activation across many visual concepts and misclassifies a fire truck as a pickup truck. SAE-FT (right) adapts by \emph{re-weighting} these existing features. It amplifies discriminative ones like ``ladder'' and reduces shared ones like ``red paint'' rather than overwriting the pre-trained feature vocabulary. This preserves the model's general knowledge while sharpening task-relevant distinctions.}
    \label{fig:intro_fig}
\end{figure}

WiSE-FT likely succeeds by balancing the zero-shot features with task-specific features; this effectively trades off ID and OOD performance. We investigate this using Sparse Autoencoders (SAEs) \cite{ng2011sparse} to achieve finer control over this balance. SAEs decompose dense representations into sparse, semantically meaningful features without assuming axis alignment \cite{cunninghamSparseAutoencodersFind2023}. While generic sparsity constraints can already limit representational drift, they offer little control over which semantic features are altered. Moreover, under standard fine-tuning, the geometry of the representation space shifts substantially, making it difficult to meaningfully compare zero-shot and fine-tuned models using a fixed SAE trained on the original representations.
\fi

To address this, we introduce Sparse Autoencoder fine-tuning (SAE-FT), a novel regularization scheme designed to prevent the destruction of semantic features during fine-tuning. We build on the linear representation hypothesis, which posits that concepts are represented as linear directions in the activation space. Standard fine-tuning often distorts these directions, degrading the model's pre-trained knowledge. SAE-FT counters this by using a Sparse Autoencoder to define the interpretable feature span of the zero-shot model. We then constrain the fine-tuning process so that any updates to the vision encoder are forced to lie within this span. This ensures that the model adapts to new tasks by re-weighting existing semantic concepts rather than overwriting them with arbitrary noise.

Our contributions are as follows:
\ifarxiv
\begin{itemize}
    \item \textbf{SAE-FT Framework}: We propose a novel fine-tuning strategy, which constrains the changes to the interpretable feature span of the pre-trained backbone. We further ensure that adaptation occurs by preserving and re-utilizing existing semantic concepts rather than overwriting them.
    \item \textbf{Performance and Efficiency}: Through extensive experiments on ImageNet and distribution-shift benchmarks, we show that SAE-FT matches or exceeds state-of-the-art robustness while avoiding text-side augmentations or injected priors. The resulting representations generalize effectively, outperforming baselines on downstream transfer tasks such as CIFAR-10 and CIFAR-100.
    \item \textbf{Mechanistic Insight}: We provide a granular analysis of feature preservation, showing that SAE-FT explicitly retains and re-weights features of the zero-shot model.
\end{itemize}
\fi

\section{Related Work}
\textbf{Robust Fine-tuning of Vision-Language Models.} A central challenge in adapting vision-language models such as CLIP \cite{radfordLearningTransferableVisual2021} is improving downstream performance while preserving robustness under distribution shifts. WiSE-FT \cite{wortsmanRobustFinetuningZeroshot2022} addresses this by interpolating the weights of the fine-tuned model with those of the zero-shot backbone, intending to regularize updates toward the pre-trained solution. Fine-tune Like You Pre-train (FLYP) \cite{goyal2022finetunelikepretrainimproved} fine-tunes CLIP using the original contrastive pre-training objective across both vision and text modalities. Subsequent approaches introduce additional constraints on fine-tuning through text-side mechanisms, such as incorporating contextual information \cite{mao2022contextawarerobustfinetuning} or injecting synthetic prompt-level features \cite{kim2025starftrobustfinetuningzeroshot}. Unlike these approaches, our method SAE-FT operates exclusively within the vision modality, achieving competitive robustness without the complexity of text-side data engineering.

\textbf{Feature Suppression and Representational Drift.} Recent work has identified a phenomenon often referred to as feature suppression or ``feature crippling,'' wherein supervised fine-tuning diminishes pre-trained features that are not directly aligned with the downstream objective \cite{mukhoti2024finetuningcripplefoundationmodel}. This representational drift has been shown to negatively affect generalization and robustness in foundation models \cite{kirkpatrick2017overcoming, kumar2022fine}. Common mitigation strategies, such as $L_2$ regularization \cite{xuhong2018explicit} or weight interpolation \cite{wortsmanRobustFinetuningZeroshot2022}, constrain parameter updates, but do not distinguish between semantically meaningful and incidental features. SAE-FT differs by explicitly identifying semantic features via dictionary learning and constraining updates with respect to this feature space during adaptation.

\textbf{Linear Representation Hypothesis and SAEs.} The Linear Representation Hypothesis (LRH) posits that high-level concepts are encoded as linear directions within a model's representation space \cite{elhage2022toy}. This serves as the theoretical basis for Sparse Autoencoders (SAEs), which decompose dense activations into an overcomplete basis of interpretable, sparse features \cite{cunninghamSparseAutoencodersFind2023}. In models such as CLIP, representations are frequently observed to be polysemantic, meaning that distinct semantic concepts are compressed into the embedding space in superposition rather than being aligned with individual orthogonal dimensions \cite{elhage2022toy}. SAEs provide a mechanism to disentangle these superposed signals into a sparse set of semantically meaningful directions. While SAEs have recently been applied to vision transformers for post-hoc mechanistic analysis \cite{lim2025patchsae, han2025causalinterpretationsparseautoencoder, joseph2025steeringclipsvisiontransformer}, they have yet to be integrated into the training loop. In this work, we shift the application of the LRH from analysis to optimization, ``exploiting'' these linear directions as a geometric constraint to prevent the crippling of foundation model features.

\section{Preliminaries}
CLIP models consist of an image encoder $f: \mathcal{X}_v \rightarrow \mathbb{R}^d$ and a text encoder $g: \mathcal{X}_t \rightarrow \mathbb{R}^d$ that map inputs from different modalities into a shared $d$-dimensional representation space. The encoders are trained using a contrastive objective, which maximizes the cosine similarity between embeddings of corresponding image-text pairs while minimizing the similarity for mismatched pairs.

\textbf{Zero-Shot Classification.} CLIP can be utilized for zero-shot classification by leveraging the semantic alignment of its joint embedding space. For a downstream classification task with $K$ classes, we transform each class label into a natural language description through prompt templating. By embedding labels into descriptive contexts (e.g., ``a photo of a \{label\}''), we align the input more closely with the natural language distribution encountered during pre-training.

Since the set of classes is typically fixed for a given task, we can pre-compute the normalized text representations for all $k \in [K]$ classes. Let $x_t^{(k)}$ denote the prompted text for class $k$. We define the class embedding $w_k$ as:
\begin{equation}
    w_k = \frac{g(x_t^{(k)})}{\|g(x_t^{(k)})\|_2} \in \mathbb{R}^d.
\end{equation}
By defining a weight matrix $W \in \mathbb{R}^{K \times d}$ where the $k$-th row corresponds to $w_k$, the classification logits for an input image $x_v \in \mathcal{X}_v$ are computed as:
\begin{equation}
    \text{logits}(x_v) = W \frac{f(x_v)}{\|f(x_v)\|_2}.
\end{equation}

\textbf{CLIP Fine-Tuning.} A common approach fine-tunes the vision encoder and a linear classification head using cross-entropy loss \cite{wortsmanRobustFinetuningZeroshot2022}. Given the classification logits defined in the zero-shot setting, fine-tuning proceeds by minimizing the cross-entropy between the predicted class probabilities and the ground-truth labels. This paradigm, adopted by methods such as WiSE-FT \cite{wortsmanRobustFinetuningZeroshot2022}, preserves the linear probing structure of CLIP while adapting the vision representations to the target task.

An alternative paradigm continues to fine-tune CLIP using the original contrastive pre-training objective over image-text pairs, updating both the vision and text encoders \cite{goyal2022finetunelikepretrainimproved}. SAE-FT operates within the linear-head, cross-entropy fine-tuning setting and introduces additional regularization on the vision representations.

\textbf{Sparse Autoencoder.} Sparse Autoencoders (SAEs) have recently emerged as a framework for mechanistic interpretability. SAEs offer a method to decompose dense, polysemantic representations into sparse, human-understandable features \cite{bricken2023towards, cunninghamSparseAutoencodersFind2023}. CLIP representations are often polysemantic, meaning semantic concepts are compressed into the embedding space in superposition rather than aligned with individual dimensions \cite{bricken2023towards, elhage2022toy}. SAEs provide a way to disentangle these superposed signals into a sparse set of semantically meaningful directions. These directions define an interpretable dictionary of features, which lets us analyze the geometry of the pre-trained representations and characterize how fine-tuning alters their structure.

Let $r \in \mathbb{R}^{d}$ be the representation of an image by the vision encoder, so $r = f(x_v)$. We train a Top-k SAE \cite{gao2024scalingevaluatingsparseautoencoders} on these representations. A Top-k SAE is a simple multi-layer perceptron that maps the representations into a sparse higher dimensional latent space ($\mathbb{R}^p$) using the TopK activation function,
\begin{align}
    s = \text{TopK}(W_e r).
\end{align}
Here $W_e \in \mathbb{R}^{p \times d}$ is the weight matrix of the SAE encoder. The training objective of the SAE is to reconstruct the representation $r$ as best as possible, given the restriction of sparsity in the higher dimensional latent space ($p > d$):
\begin{align}
    \tilde{r} = W_d s.
\end{align}
The decoder weights $W_d \in \mathbb{R}^{d \times p}$ therefore define a dictionary that maps sparse feature activations $s_1, \dots, s_{p}$ to directions in the CLIP representation space.

\section{Representational Drift in CLIP Fine-Tuning}\label{sec: drift}
Before introducing SAE-FT, we analyze how standard and robust fine-tuning procedures alter the internal geometry of CLIP vision representations. This analysis reveals systematic representational drift that limits both interpretability and robustness, and directly motivates the geometric constraints introduced in Section~\ref{sec:sae_ft}.

We compare the Centered Kernel Alignment (CKA) similarity \cite{kornblithSimilarityNeuralNetwork2019a} between the representations of the zero-shot model, a standard fine-tuned model, and a robust fine-tuned model. We choose WiSE-FT as the robust fine-tuning method because it only uses the vision encoder and indirectly regularizes the visual representations.

\begin{figure}[t]
  \centering
  \includegraphics[width=0.6\textwidth]{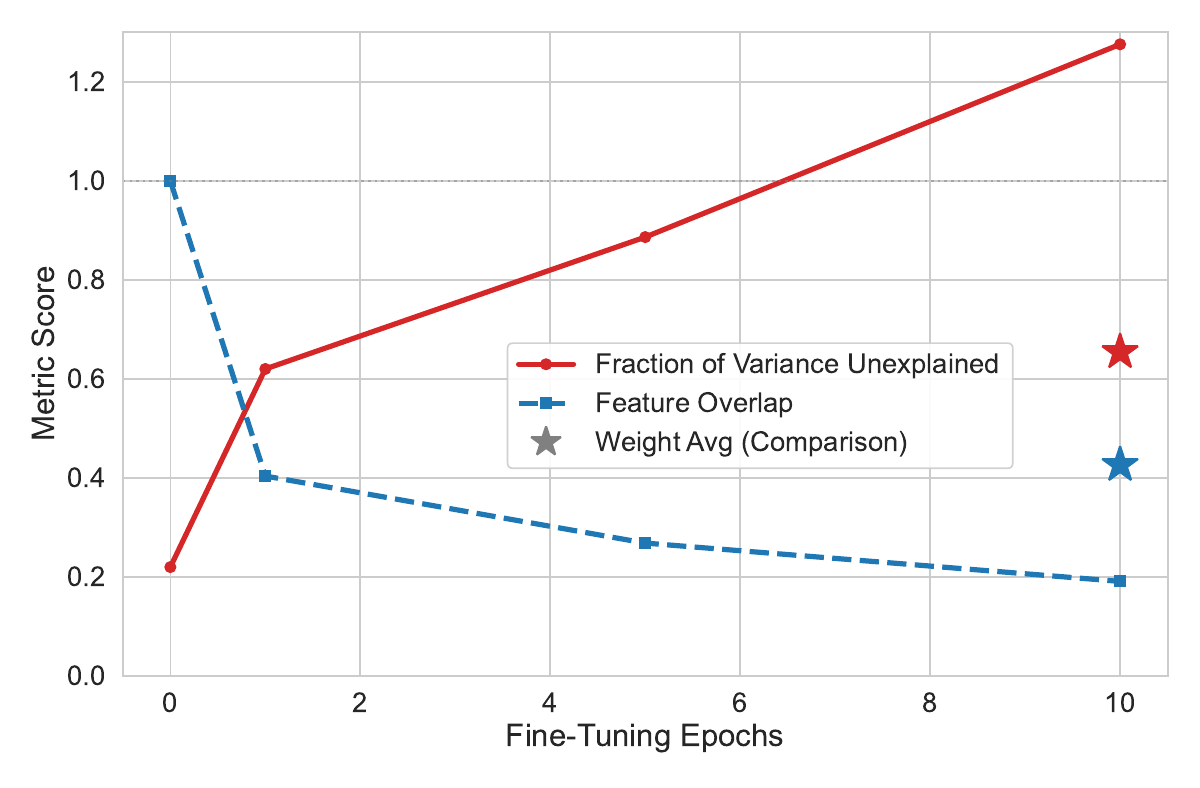}
  \caption{Standard fine-tuning causes the original dictionary to collapse (Fraction of Variance Unexplained, FVU $> 1.0$) and erases $\sim80\%$ of semantic concepts (Feature Overlap). While Weight Averaging improves the stability of the learned dictionary it fails to prevent significant feature replacement.}
  \label{fig: SAE}
\end{figure}

\ifarxiv
\begin{table}[t]
\centering
\small
\setlength{\tabcolsep}{3pt}
\caption{CKA similarity matrix for vision encoder representations. Rep. avg. denotes the element-wise average of the Zero-shot and Fine-tuned activation vectors. Fine-tuning completely changes the representations. WiSE-FT recovers the geometry of the zero-shot better than direct representation averaging.}
\begin{tabularx}{\columnwidth}{l Y Y Y Y}
\toprule
& Zero-shot & Fine-tuned & Rep. avg. & WiSE-FT \\
\midrule
Zero-shot  & 1.00 & 0.40 & 0.67 & 0.83 \\
Fine-tuned &      & 1.00 & 0.82 & 0.59 \\
Rep. avg.  &      &      & 1.00 & 0.93 \\
WiSE-FT    &      &      &      & 1.00 \\
\bottomrule
\end{tabularx}
\label{tab: cka comparison}
\end{table}

Fine-tuning fundamentally alters the internal representations of the model; this shift can be partially reversed through weight-space averaging. Table \ref{tab: cka comparison} shows that the CKA similarity between fine-tuned and zero-shot models drops to 0.40, which confirms major representational changes. Comparing weight-space interpolation (WiSE-FT) with direct representation averaging (Rep. Avg.) shows that, although both methods combine information from the two models, WiSE-FT produces representations substantially closer to the zero-shot geometry (0.83 similarity) than representation averaging (0.67 similarity). This indicates that weight-space interpolation preserves the pre-trained model geometry, whereas output interpolation remains largely dominated by the drifted fine-tuned structure.

Further we compare the representations of the fine-tuned model to the zero-shot model with an SAE. The SAE is trained on the zero-shot model and used for all models. Figure \ref{fig: SAE} shows the Fraction of Variance that is unexplained (FVU) by the SAE for different fine-tuning epochs and the weight averaged model. It also shows the percentage of SAE features of the zero-shot model that are preserved when applying the SAE to other models.

The analysis shows that the pre-trained dictionary effectively collapses when applied to fine-tuned representations. Standard fine-tuning results in an FVU $> 1.0$, implying that the feature space has drifted so severely that the original dictionary performs worse than a zero-vector baseline. This confirms that fine-tuning does not merely adjust feature activations but fundamentally alters the basis of the representation space. Even with the regularization provided by WiSE-FT, only $43\%$ of the original features are preserved, and the high FVU ($0.65$) indicates that the resulting representations remain difficult to interpret using the original vocabulary.

\fi

These findings demonstrate that geometric drift limits the interpretability and robustness of standard fine-tuning. A simple method to limit geometric drift is to regularize the representations of the fine-tuned model with the representations of the pre-trained model. Let $r^0$ and $r^{ft}$ be the representations of the pre-trained and fine-tuned model respectively and let $\Delta r := r^{ft} - r^{0}$ be the difference in representations. The following regularization is added to the standard cross-entropy loss of fine-tuning:
\begin{align}
    \mathcal{L} = \lambda ||\Delta r||_2^2.
\end{align}
We note that this regularization is similar to the LDIFS method introduced by \citet{mukhoti2024finetuningcripplefoundationmodel}. However the regularization is only applied to the final vision representations in our case.

\begin{table}[t]
\centering
\small
\setlength{\tabcolsep}{3pt}
\caption{Representation and feature comparison between zero-shot, $L_2$ regularized, standard fine-tuning and WiSE-FT models. $L_2$ regularization restricts the geometric changes to the model.}
\begin{tabularx}{\columnwidth}{l Y Y Y Y}
\toprule
& CKA with zero-shot & FVU of SAE & Feature overlap & Feature Entropy \\
\midrule
Zero-shot  & 1.00 & 0.22 & 1.00 & 2.63 \\
FT & 0.40 & 1.28 & 0.19 & 2.67 \\
WiSE-FT & 0.83 & 0.65 & 0.43 & 2.66 \\
$L_2$ reg &  1.00 & 0.23 & 0.67 & 2.60 \\
\bottomrule
\end{tabularx}
\label{tab: l2_comp}
\end{table}

$L_2$ regularization limits the geometric drift, but features can still change. Table \ref{tab: l2_comp} shows the CKA with the zero-shot and the FVU of the SAE trained on the pre-trained model for the $L_2$ regularized model. The CKA is at $1.0$ showing that the regularization limits geometric drift and the representations of the fine-tuned model are almost equal to the representations of the pre-trained model up to isotropic scaling and orthogonal projections. The FVU of the SAE also stays low, allowing us to compare the features of the fine-tuned to the zero-shot model. \ifarxiv The entropy of the features largely does not change. This means that the relative importance of features does not shift to a few dominant features. However the \fi feature overlap of the $L_2$ regularized model and pre-trained model is relatively low at $0.67$. This shows that the model adapts the features it uses. $L_2$ regularization only yields control over the overall change in geometry, but not over the more specific feature adaptation.

This motivates our proposed SAE-FT framework, which explicitly constrains the optimization trajectory to stay within the valid geometric span of the zero-shot SAE. This does not only limit the geometric drift during fine-tuning, but also yields control over how features change, resulting in a more informed, interpretable and flexible fine-tuning method. While agnostic regularization methods such as $L_2$ already recover most of the robustness by preserving the overall geometry, they treat all directions equally and cannot distinguish between semantically meaningful and incidental changes. SAE-FT addresses this gap by operating in a learned feature basis.

\section{SAE-FT} \label{sec:sae_ft}
\begin{figure}[t]
    \centering
    \includegraphics[width=0.8\columnwidth]{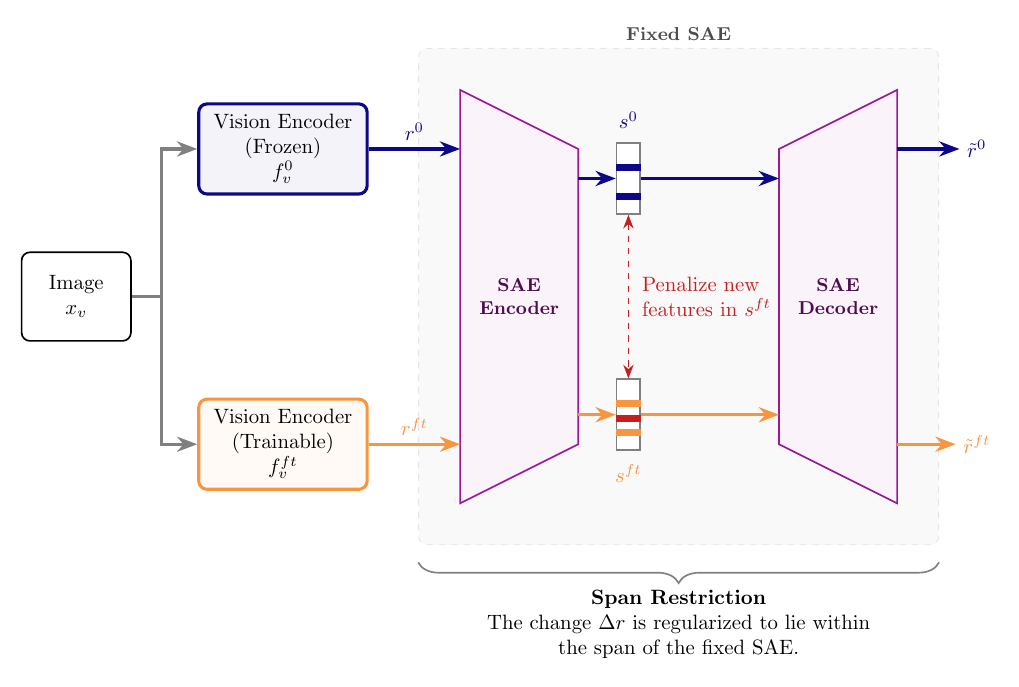}
    \caption{Schematic overview of SAE-FT. Changes compared to the zero-shot model are encouraged to remain inside the span of the fixed SAE and the addition of new SAE features is penalized.}
    \label{fig:overview}
\end{figure}

\ifarxiv
\begin{algorithm}[t]
\caption{SAE-FT Training and Inference}
\label{alg:sae-ft}
\begin{algorithmic}[1]
    \STATE \textbf{Unsupervised SAE Training}
    \STATE \textit{Requires no labels or test data.}
    \STATE Train SAE on zero-shot representations $r^0$ from $f_0$ until convergence.
    \STATE Freeze SAE weights.
    \STATE \textbf{Task-Specific Fine-tuning}
    \STATE Initialize $f \gets f_0$ and classifier $W$
    \FOR{each minibatch $(x_v, y)$}
        \STATE $r^{ft} \gets f(x_v)$, $r^{0} \gets f_0(x_v)$
        \STATE $s^{0} \gets \mathrm{SAE}_{\text{enc}}(r^{0})$ \textbf{(no grad)}
        \STATE $s^{ft} \gets \mathrm{SAE}_{\text{enc}}(r^{ft})$
        \STATE $\text{logits} \gets W r^{ft}$
        \STATE $\mathcal{L} \gets \mathcal{L}_{CE} + \lambda \mathcal{L}_{\text{add}}$
        \STATE Update $f, W$
    \ENDFOR
    \STATE \textbf{Inference:} $\hat{y} = \arg\max W f(x_v)$
\end{algorithmic}
\end{algorithm}
\fi

The goal of our regularization is to constrain fine-tuning to the interpretable features of the pre-trained model. Specifically, we enforce that all changes to the representations can be explained by the zero-shot SAE, and we explicitly restrict which features are allowed to vary. This ensures that the general geometry of the representation space is preserved, while allowing us to penalize specific semantic shifts, such as the emergence of spurious features.

Let $r^{ft}, r^0 \in \mathbb{R}^{d}$ be the representations of the fine-tuned and zero-shot model respectively. We utilize a pre-trained Sparse Autoencoder with an encoder $\text{SAE}_{enc}:\mathbb{R}^{d} \to \mathbb{R}^{p}$ and a linear decoder $W_d \in \mathbb{R}^{d \times p}$. Let $s^0 := \text{SAE}_{enc}(r^0)$ and $s^{ft} := \text{SAE}_{enc}(r^{ft})$ denote the sparse feature activations. We define the change in feature space as $\Delta s := s^{ft} - s^0$ and the change in representation space as $\Delta r:= r^{ft} - r^0$.

To ensure that representational updates remain within the semantic span of the dictionary, we introduce a residual alignment penalty:
\begin{align}
    \mathcal{L}_{\text{resid}} := ||\Delta r - W_d(\Delta s)||_2^2.
\end{align}
This term minimizes the component of $\Delta r$ that is orthogonal to the decoder's span, forcing the fine-tuning updates to be expressible as a linear combination of interpretable features.\ifarxiv{} Figure \ref{fig: residual} shows a visualization of this loss term.\fi

We propose two regularization strategies to control feature drift:

\textbf{1. Sparse Feature Regularization.} A naive approach is to simply enforce sparsity on the feature differences, encouraging the model to change as few features as possible:
\begin{align}
    \mathcal{L}_{\text{sparse}} := \lambda_{\text{resid}} \mathcal{L}_{\text{resid}} + \lambda_{\text{sparse}} ||\Delta s||_1.
\end{align}

\textbf{2. Feature Preservation.} Pre-trained CLIP models capture a vast range of concepts, many of which are irrelevant to a specific downstream task. Rather than preserving all features equally, we hypothesize that robust fine-tuning should focus on re-weighting relevant features while preventing the addition of new, task-irrelevant concepts. We achieve this by penalizing the activation of features that were inactive in the zero-shot model:
\begin{align}
    m_k &:= \mathbb{I}(s^0_k \neq 0) \\
    \mathcal{L}_{\text{add}} &:= \lambda_{\text{resid}} \mathcal{L}_{\text{resid}} + \lambda_{\text{add}} \frac{1}{p} \sum_{k=1}^{p} (1 - m_k) | s^{ft}_k |. \label{equ: l_add}
\end{align}
When using a Top-K SAE (where the number of active features is fixed), this penalty implicitly acts as a strict support-set constraint. Since the model must maintain $K$ active features, penalizing the addition of new features ($m_k=0$) forces the model to rely solely on re-weighting the original features ($m_k=1$), effectively locking the semantic support of the model.\ifarxiv{} Figure \ref{fig: add_penalty} shows how this penalizes feature change.\fi

\ifarxiv
\begin{figure*}[tbp]
    \centering
    \begin{minipage}[t]{0.48\textwidth}
        \centering
        \vbox to 5.5cm{\vfill
        \begin{tikzpicture}[scale=0.8, >=Stealth, thick]
            \draw[->, gray!40] (0,0) -- (2.5,0);
            \draw[->, gray!40] (0,0) -- (0,2.5);
            \draw[->, gray!40] (0,0) -- (-1.2,-1.2);
            \draw[->, pBlue] (0,0) -- (1.8,1.2) node[right, font=\tiny] {$r^0$};
            \draw[->, pOrange] (0,0) -- (0.6,2.0) node[above, font=\tiny] {$r^{ft}$};
            \draw[<->, pPurple, dashed, line width=0.8pt] (1.8,1.2) -- (0.6,2.0) node[midway, above right, font=\tiny] {$\Delta r$};
            \draw[->, pBlue!70] (0,0) -- (1.8,-0.5) node[right, font=\tiny] {$\tilde{r}^0$};
            \draw[->, pOrange!70] (0,0) -- (0.6,-1.1) node[below, font=\tiny] {$\tilde{r}^{ft}$};
            \draw[<->, pPurple, dashed, line width=0.8pt] (1.8,-0.5) -- (0.6,-1.1) node[midway, below right, font=\tiny] {$\Delta \tilde{r}$};
            \node[text=pRed, font=\bfseries\tiny, yshift=-1.5cm] at (1,0) {$\Delta r \approx \Delta \tilde{r}$};
        \end{tikzpicture}
        \vfill}
        \caption{Visualization of the residual loss. It enforces the change in representations and reconstructed representations to be similar.}
        \label{fig: residual}
    \end{minipage}
    \hfill
    \begin{minipage}[t]{0.48\textwidth}
        \centering
        \vbox to 5.5cm{\vfill
        \begin{tikzpicture}[scale=0.5]
            \draw[->] (0,0) -- (0,4.5) node[rotate=90, midway, above, font=\tiny] {magnitude};
            \draw[->] (0,0) -- (4,0) node[midway, below, font=\tiny] {features};
            \node[font=\small] at (2,4.8) {$s^{ft}$};
            \fill[pOrange!40, draw=black, thin] (0.2,0) rectangle (0.6,3.8);
            \fill[pOrange!40, draw=black, thin] (0.8,0) rectangle (1.2,0.8);
            \fill[pRed, draw=black, thin] (1.4,0) rectangle (1.8,2.2);
            \fill[pOrange!40, draw=black, thin] (2.0,0) rectangle (2.4,0.4);
            \fill[pOrange!40, draw=black, thin] (2.6,0) rectangle (3.0,2.5);
            \begin{scope}[xshift=6cm]
                \draw[->] (0,0) -- (0,4.5) node[rotate=90, midway, above, font=\tiny] {magnitude};
                \draw[->] (0,0) -- (4,0) node[midway, below, font=\tiny] {features};
                \node[font=\small] at (2,4.8) {$s^{0}$};
                \fill[pBlue!40, draw=black, thin] (0.2,0) rectangle (0.6,2.5);
                \fill[pBlue!40, draw=black, thin] (0.8,0) rectangle (1.2,1.8);
                \fill[pBlue!40, draw=black, thin] (2.0,0) rectangle (2.4,1.2);
                \fill[pBlue!40, draw=black, thin] (2.6,0) rectangle (3.0,3.8);
                \fill[pBlue!40, draw=black, thin] (3.2,0) rectangle (3.6, 1);
            \end{scope}
        \end{tikzpicture}
        \vfill}
        \caption{Visualization of the feature preservation regularization. New features are penalized (dark orange), while changing the magnitude of existing features is not penalized.}
        \label{fig: add_penalty}
    \end{minipage}
\end{figure*}
\fi

SAE-FT does not update the SAE during fine-tuning and does not employ the SAE during inference, keeping computational overhead limited. \ifarxiv As shown in Algorithm \ref{alg:sae-ft} once\fi{} the SAE is trained on the frozen representations of the zero-shot model, it is only used to compute the regularization terms, without any update to its parameters.

\subsection{SAE-FT vs. Standard Regularization}
\ifarxiv
A standard approach to prevent drift is to apply $L_1$ regularization directly to the representation differences:
\begin{align}\mathcal{L}_{\text{std}} = \lambda ||\Delta r||_1 = \lambda \sum_{i=1}^{d} |\Delta r_i|.\end{align}
This penalty assumes that the representation basis vectors (the neurons) are the fundamental units of meaning (axis alignment). However, in dense models like CLIP, features are often polysemantic and stored in superposition, meaning that individual neurons $r_i$ do not correspond to distinct concepts. Minimizing $\mathcal{L}_{std}$ therefore restricts changes along arbitrary, non-semantic axes.

In contrast, SAE-FT applies sparsity in the feature space:
\begin{align}\mathcal{L}_{\text{SAE}} \propto ||\Delta s||_1 = \sum_{k=1}^{p} |\Delta s_k|.\end{align}
By regularizing $\Delta s$, we apply constraints along the directions of the learned dictionary $W_d$. Unlike the standard basis, these directions are optimized to be semantically distinct. Thus, SAE-FT regularizes the model's semantic content directly, allowing for significant changes in the raw activation space ($\Delta r$) as long as they correspond to limited updates in the feature space.

In contrast to $L_1$ regularization and SAE-FT, $L_2$ regularization is invariant to certain directions in the representation space. This results in a regularization that regularizes geometric drift, but gives no control over specific feature change.
\fi

\section{Experiments}\label{sec: experiments}
We conduct experiments to show the robustness and generalization capabilities of models fine-tuned with SAE-FT. We compare SAE-FT to state-of-the-art methods on distribution shifts, specific OOD datasets and zero-shot generalization to downstream datasets.

\subsection{Experimental Setup}\label{sec: exp_setup}
We evaluate SAE-FT against several robust fine-tuning methods under three evaluation settings. Sections~\ref{sec: exp_IN} and~\ref{sec: exp_gen} consider models fine-tuned on the ImageNet training dataset \cite{5206848}. Section~\ref{sec: exp_IN} evaluates performance on ImageNet (IN) and standard distribution shift benchmarks, including ImageNet-R (IN-R) \cite{hendrycksManyFacesRobustness2021}, ImageNet-A (IN-A) \cite{hendrycksNaturalAdversarialExamples2021}, ImageNet-Sketch (IN-S) \cite{wang2019learning}, and ImageNet-V2 (IN-V2) \cite{rechtImageNetClassifiersGeneralize2019}. Section~\ref{sec: exp_gen} assesses generalization by evaluating ImageNet-fine-tuned models on additional downstream datasets without further task-specific fine-tuning. Section~\ref{sec: exp_wilds} evaluates robustness on iWilds benchmarks, where models are fine-tuned and evaluated separately for each dataset.

\ifarxiv
\textbf{WiSE-FT} \cite{wortsmanRobustFinetuningZeroshot2022} averages the parameters of a linear-head fine-tuned vision model with the zero-shot model, encouraging updates to remain close to the pre-trained weights. Only the vision encoder and linear head are fine-tuned.

\textbf{Context-Aware Robust Fine-Tuning (CAR-FT)} \cite{mao2022contextawarerobustfinetuning} regularizes the vision encoder to retain context understanding by matching context distributions from the frozen text encoder prompted with pre-defined templates.

\textbf{Fine-tune Like You Pre-train (FLYP)} \cite{goyal2022finetunelikepretrainimproved} fine-tunes the entire vision-language model by continuing to optimize the original contrastive pretraining loss on downstream labeled data. It casts class labels as text prompts and updates both the vision and text encoders under the contrastive objective, aligning fine-tuning more closely with how the model was pretrained.

\textbf{Calibrated Robust Fine-Tuning (CaRot)} \cite{oh2024towards} applies a self-distillation strategy where the model is trained to match the predictions of an exponential moving average (EMA) of its own weights, alongside the contrastive FLYP objective. Both the vision and text encoders are fine-tuned.

\textbf{Spurious Textual Alignment Regularization (StarFT)} \cite{kim2025starftrobustfinetuningzeroshot} fine-tunes both vision and text encoders using the FLYP objective. It aligns predictions on text prompts with injected spuriosity features to the zero-shot teacher, requiring external LLM-generated textual data.

\textbf{SAE-FT} (ours) operates on the vision encoder only, using the cross-entropy linear-head fine-tuning framework. We add a regularization term that constrains updates to the interpretable semantic span of a pre-trained Sparse Autoencoder, explicitly restricting which features can change (Equation~\ref{equ: l_add}). Hyperparameters $\lambda_{\mathrm{res}}$ and $\lambda_{\mathrm{add}}$ are chosen via search to balance the residual and feature-addition penalties. All results for our method are averaged over three training runs and we report standard deviations as subscripts.
\fi

The results for standard fine-tuning (FT), FLYP, CAR-FT, CaROT and StarFT are taken from \citet{kim2025starftrobustfinetuningzeroshot}. The results for WiSE-FT in Section \ref{sec: exp_IN} are taken from \citet{wortsmanRobustFinetuningZeroshot2022}.

\subsection{Robust fine-tuning on ImageNet and distribution shifts} \label{sec: exp_IN}
Table~\ref{tab: ViT-B/16} reports results for the OpenAI ViT-B/16 model evaluated on ImageNet and its distribution shifts. SAE-FT achieves competitive in-distribution performance, reaching the second-highest ImageNet accuracy ($82.9\%$), while attaining the highest average accuracy across distribution shift benchmarks.\ifarxiv{} Compared to other vision-encoder-only methods, SAE-FT improves ImageNet accuracy by $1.0$ percentage point over CAR-FT and improves average out-of-distribution performance by $0.2$ percentage points over WiSE-FT.\fi

\begin{table}[t]
\centering\small\setlength{\tabcolsep}{3pt}
\caption{Robust fine-tuning results on ImageNet and distribution shift benchmarks for the OpenAI ViT-B/16 model. SAE-FT matches state of the art methods and improves over other methods that only fine-tune the vision encoder.}
\begin{tabularx}{\columnwidth}{l Y Y Y Y Y Y}
\toprule
\textbf{Method} & \multicolumn{6}{c}{\textbf{ViT-B/16}} \\
\cmidrule(lr){2-7}
& IN & IN-R & IN-A & IN-S & IN-V2 & Avg. \\
\midrule
Zero-shot & 68.3 & 77.7 & 50.0 & 48.3 & 61.9 & 59.5 \\
FT & 81.3 & 71.3 & 44.5 & 49.1 & 71.7 & 59.1 \\
FLYP & 82.6 & 71.4 & 48.5 & 49.8 & 72.7 & 60.6 \\
CAR-FT & 81.9 & 75.6 & 50.0 & 51.5 & 72.8 & 62.5 \\
CaRot & \textbf{83.1} & 76.2 & 51.3 & 51.9 & \textbf{74.3} & 63.7 \\
WiSE-FT & 81.7 & \textbf{78.7} & 52.2 & \textbf{53.9} & 72.8 & 64.4 \\
StarFT & 82.9 & 77.7 & \textbf{53.7} & 52.5 & 73.8 & 64.4 \\
\midrule
\textbf{SAE-FT} &
\makecell{82.9 \\ {\small $\pm 0.1$}} &
\makecell{78.5 \\ {\small $\pm 0.1$}} &
\makecell{52.6 \\ {\small $\pm 0.4$}} &
\makecell{53.4 \\ {\small $\pm 0.0$}} &
\makecell{73.9 \\ {\small $\pm 0.1$}} &
\makecell{\textbf{64.6} \\ {\small $\pm 0.1$}} \\
\bottomrule
\end{tabularx}
\label{tab: ViT-B/16}
\end{table}

\subsection{Transfer and generalization to downstream datasets}\label{sec: exp_gen}
\ifarxiv
To evaluate whether the regularized representations learned by SAE-FT generalize beyond ImageNet, we additionally evaluate all methods on a set of downstream classification benchmarks, including CIFAR-10, CIFAR-100 \cite{krizhevsky2009learning}, Caltech-101 \cite{fei2007learning}, and STL-10 \cite{coates2011stl10}. We follow the standard CLIP transfer evaluation protocol used in prior work \cite{kim2025starftrobustfinetuningzeroshot}, in which the fine-tuned ImageNet model is evaluated on downstream datasets without additional task-specific fine-tuning.

Concretely, for each downstream dataset, images are passed through the fine-tuned vision encoder to obtain visual representations, which are then classified using the original zero-shot CLIP text classifier constructed from dataset-specific class names.
\fi

Results for ViT-B/16 are shown in Table~\ref{tab: generalization}. SAE-FT achieves the highest average transfer accuracy across all evaluated datasets, outperforming both standard fine-tuning and existing robust fine-tuning methods.\ifarxiv{} These results indicate that constraining representation updates via a fixed, interpretable feature space not only preserves robustness to distribution shifts, but also yields representations that transfer effectively to diverse downstream tasks.\fi

\begin{table}[t]
\centering\small\setlength{\tabcolsep}{3pt}
\caption{Generalization performance of ViT-B/16 models fine-tuned on ImageNet and evaluated on downstream transfer benchmarks. SAE-FT representations generalize better than other representations from robust fine-tuning methods.}
\begin{tabularx}{\columnwidth}{l Y Y Y Y Y}
\toprule
\textbf{Method} & \multicolumn{5}{c}{\textbf{ViT-B/16}} \\
\cmidrule(lr){2-6}
& C-10 & C-100 & Cal101 & STL10 & Avg. \\
\midrule
Zero-shot & 90.8 & 68.2 & 89.6 & 98.3 & 86.7 \\
FT & 87.7 & 63.6 & 85.7 & 95.3 & 83.1 \\
FLYP & 90.0 & 64.2 & 87.4 & 98.5 & 85.0 \\
CAR-FT & 89.7 & 65.9 & 88.2 & 96.7 & 85.2 \\
CaRot & 91.1 & 66.7 & 89.0 & 98.7 & 86.5 \\
WiSE-FT & 91.2 & 69.6 & 87.6 & 98.1 & 86.5 \\
StarFT & 91.4 & 69.0 & \textbf{89.7} & \textbf{99.0} & 87.3 \\
\midrule
\textbf{SAE-FT} &
\makecell{\textbf{91.9} \\ {\small $\pm 0.2$}} &
\makecell{\textbf{71.2} \\ {\small $\pm 0.4$}} &
\makecell{89.5 \\ {\small $\pm 0.1$}} &
\makecell{98.7 \\ {\small $\pm 0.0$}} &
\makecell{\textbf{87.8} \\ {\small $\pm 0.2$}} \\
\bottomrule
\end{tabularx}
\label{tab: generalization}
\end{table}

\subsection{iWilds datasets}\label{sec: exp_wilds}
We evaluate our approach on two challenging datasets from the iWilds benchmark \cite{wilds2021}: iWildCam and Feature Map of the World (FMoW). \ifarxiv Both datasets are designed to assess robustness under real-world distribution shifts, making them well suited for studying OOD generalization. \fi For iWildCam, we report the macro-averaged F1 score; for FMoW, we report accuracy on the ID test split and worst-group accuracy on the OOD test split.

Table \ref{tab: iWilds} summarizes the results for the OpenAI ViT-B/16 CLIP model. \ifarxiv Zero-shot performance is consistently low, highlighting the difficulty of both datasets. Fine-tuning yields substantial gains, and recent methods such as FLYP, CAR-FT, and StarFT further improve ID performance. WiSE-FT shows strong ID results when the interpolation parameter is optimally tuned ($\alpha=0.8$ for iWildCam and $\alpha=0.9$ for FMoW), though its OOD improvements are more limited. \fi Across both datasets, SAE-FT achieves the best or near-best performance. On iWildCam, it attains the highest OOD macro F1 score while remaining competitive on ID data. On FMoW, SAE-FT matches the best ID accuracy and achieves the strongest OOD worst-group accuracy.\ifarxiv{} We additionally include the $L_2$ baseline; SAE-FT matches $L_2$ on FMoW and outperforms it on iWildCam, where the zero-shot model is weak and more adaptation is needed.\fi

\begin{table}[t]
\centering\small\setlength{\tabcolsep}{3pt}
\caption{Results for the ViT-B/16 model on iWildCam and FMoW. For iWildCam we report the f1-macro score and for FMoW the accuracy for the ID test set and the worst group accuracy for the OOD test set. For WiSE-FT the optimal $\alpha$ is 0.8 for iWildCam and 0.9 for FMoW.}
\label{tab: iWilds}
\begin{tabularx}{\columnwidth}{l Y Y Y Y}
\toprule
\textbf{Method} & \multicolumn{2}{c}{\textbf{iWildCam}} & \multicolumn{2}{c}{\textbf{FMoW}} \\
\cmidrule(lr){2-3} \cmidrule(lr){4-5}
& ID & OOD & ID & OOD \\
\midrule
Zero-shot & 8.7 & 11.0 & 20.4 & 18.7 \\
FT & 47.2 & 35.6 & 68.6 & 40.2 \\
FLYP & 48.5 & 36.6 & 68.6 & 40.1 \\
CAR-FT & 45.8 & 37.0 & 68.4 & 40.7 \\
CaRot & 40.6 & 29.2 & 51.9 & 26.8 \\
WiSE-FT (0.5) & 38.6 & 30.8 & 61.5 & 40.3 \\
WiSE-FT (opt.) & 44.8 & 33.1 & \textbf{69.2} & 42.1 \\
StarFT & \textbf{50.1} & 37.1 & 68.4 & 41.0 \\
$L_2$ & 48.1 & 34.7 & \textbf{69.2} & \textbf{42.8} \\
\midrule
\textbf{SAE-FT} &
\makecell{49.6 \\ {\small $\pm 0.2$}} &
\makecell{\textbf{38.1} \\ {\small $\pm 1.6$}} &
\makecell{\textbf{69.2} \\ {\small $\pm 0.2$}} &
\makecell{\textbf{42.8} \\ {\small $\pm 0.5$}} \\
\bottomrule
\end{tabularx}
\end{table}

\subsection{Comparison to representation regularization baselines}\label{sec: exp_base}
\ifarxiv
To better understand the role of the SAE-based regularization, we compare SAE-FT to several generic alternatives that constrain changes in the vision representations. Specifically, we consider $L_1$ and $L_2$ penalties on representation differences, as well as a PCA-based regularization that restricts updates to a fixed low-dimensional subspace learned from the zero-shot model. These baselines are evaluated on ViT-B/16 and are summarized in Table~\ref{tab: reg_baselines}.
\fi

\begin{table}[t]
\centering\small\setlength{\tabcolsep}{3pt}
\caption{Comparison of SAE-FT to representation regularization baselines for ViT-B/16. We report $L_1$, $L_2$, and PCA-based constraints applied to representation changes during fine-tuning. SAE-FT yields a slight performance improvement.}
\begin{tabularx}{\columnwidth}{l Y Y Y Y Y Y}
\toprule
\textbf{Method} & \multicolumn{6}{c}{\textbf{ViT-B/16}} \\
\cmidrule(lr){2-7}
 & IN & IN-R & IN-A & IN-S & IN-V2 & Avg. \\
\midrule
$L_1$ & 82.8 & 78.2 & 52.2 & 53.0 & 73.6 & 64.3 \\
$L_2$ & 82.8 & 78.6 & 52.4 & 53.1 & \textbf{73.9} & 64.5 \\
PCA & 82.6 & \textbf{78.7} & 52.4 & 53.1 & 73.6 & 64.5 \\
\midrule
\textbf{SAE-FT} &
\makecell{\textbf{82.9} \\ {\small $\pm 0.1$}} &
\makecell{78.5 \\ {\small $\pm 0.1$}} &
\makecell{\textbf{52.6} \\ {\small $\pm 0.4$}} &
\makecell{\textbf{53.4} \\ {\small $\pm 0.0$}} &
\makecell{\textbf{73.9} \\ {\small $\pm 0.1$}} &
\makecell{\textbf{64.6} \\ {\small $\pm 0.1$}} \\
\bottomrule
\end{tabularx}
\label{tab: reg_baselines}
\end{table}

Geometric drift is a key reason for the degradation of robustness during fine-tuning. \ifarxiv Limiting it results in a fine-tuned model that performs better in-distribution and on distribution shifts. $L_1$, $L_2$, and PCA-based regularization all substantially reduce the degradation in robustness compared to unregularized fine-tuning, their performance is broadly similar across evaluated benchmarks. SAE-FT achieves slightly higher average accuracy, but its primary advantage lies not in large performance gains, rather in the structure of the imposed constraint. Unlike generic regularization applied in the raw activation space, SAE-FT enforces sparsity and preservation in a learned, semantically meaningful feature basis derived from the zero-shot model. This enables controlled and interpretable modification of representations during fine-tuning, while maintaining competitive robustness and in-distribution performance.\fi

\section{Analysis of SAE-FT} \label{sec: analysis}
In this section, we use the SAE employed for regularization to analyze the fine-tuning process. This analysis is performed using the ViT-B/16 model and the ImageNet test set. We evaluate SAE-FT with the feature-addition regularization described in Equation~\ref{equ: l_add}.

\begin{figure*}[t]
    \centering
    \includegraphics[width=0.8\linewidth]{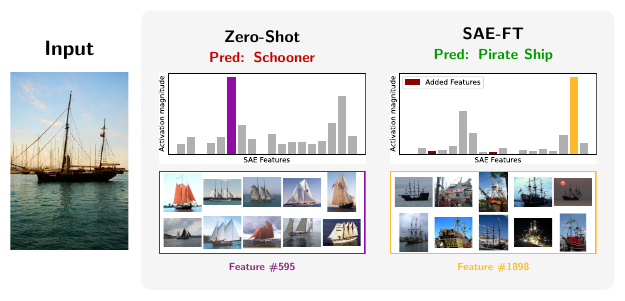}
    \caption{Feature re-weighting in SAE-FT. We analyze an image of a pirate ship misclassified by the zero-shot model as a schooner. SAE-FT corrects the prediction by amplifying the task-relevant ``pirate ship'' feature while retaining the ``schooner'' feature with reduced importance.}
    \label{fig:feature_shift_ship}
\end{figure*}

\subsection{Feature Statistics}
\ifarxiv
To better understand how fine-tuning changes the features during regularized fine-tuning, we compare overall statistics of the SAE features of the zero-shot, $L_2$ regularized and SAE regularized models in Table \ref{tab: feature_comp}.

For both regularization methods the representations remain very similar to the representations of the zero-shot model (CKA $\sim 1.0$). This leads to an accurate reconstruction by the zero-shot model's SAE for both fine-tuned models (FVU $\leq 0.25$). The feature overlap and entropy of the features shows how SAE-FT behaves differently from standard norm regularization. Because of the specific penalization of removing and adding features, SAE-FT has a higher feature overlap with the zero-shot model than $L_2$ regularization. The lower feature entropy of SAE-FT reveals that while SAE-FT retains most of the features, it re-weights the features and gives more importance towards task-specific features. As $L_2$ regularization is invariant to feature directions, it does not allow for feature re-weighting and its feature entropy remains higher. This increased flexibility in feature usage gives SAE-FT the ability to yield strong performance on datasets which are hard for the zero-shot and subsequently the $L_2$ regularized model (see results for iWildCam in Table \ref{tab: iWilds} and Appendix \ref{apdx: comp_l2}).
\fi

\begin{table}[t]
\centering\small\setlength{\tabcolsep}{3pt}
\caption{Representation and feature comparison between zero-shot, $L_2$ regularized and SAE-FT models. Both $L_2$ regularization and SAE-FT restrict the geometric changes to the model, but SAE-FT re-uses more features, while re-weighting them.}
\begin{tabularx}{\columnwidth}{l Y Y Y Y}
\toprule
& CKA with zero-shot & FVU of SAE & Feature overlap & Feature Entropy \\
\midrule
Zero-shot  & 1.00 & 0.22 & 1.00 & 2.63 \\
$L_2$ reg &  1.00 & 0.23 & 0.67 & 2.60 \\
SAE-FT  &    0.99  & 0.25 & 0.78 & 2.36 \\
\bottomrule
\end{tabularx}
\label{tab: feature_comp}
\end{table}

\subsection{Qualitative Analysis of Feature Re-weighting}
To examine these changes concretely, we apply the SAE to individual samples, specifically focusing on images misclassified by the zero-shot model but correctly classified by SAE-FT. Figure~\ref{fig:feature_shift_ship} illustrates an example of a pirate ship, initially misclassified as a schooner. Comparing the activation patterns reveals a clear shift in feature priority. In the zero-shot model, feature 595 (associated with schooners) has the highest activation, while feature 1898 (associated with pirate ships) is secondary. In the SAE-FT model, this ranking is flipped. The fine-tuned model does not erase the concept of the schooner, which is visually present, but rather up-weights the specific attributes that distinguish the pirate ship, which are more critical for the classification task. We observe this ``feature re-weighting'' consistently across various classes, additional qualitative examples are provided in Appendix \ref{apdx:res_qual}.

\ifarxiv
\subsection{Mechanism of Improvement}
These findings suggest that SAE-FT succeeds because it forces the model to focus on a sparser, more relevant subset of features rather than learning new representations from scratch. The large-scale pre-training of the zero-shot model captures a vast spectrum of visual concepts, however, downstream classification often requires only the subset of features most aligned with the target classes. Standard fine-tuning fundamentally changes the representations to prioritize task-specific features, which deteriorates the model's robustness and ability to generalize to other tasks. Norm regularized fine-tuning preserves the representation geometry, but does not distinguish between necessary semantic shifts and overall geometric drift. SAE-FT constrains the geometric changes, preserving the general features of the model, but simultaneously allows the model to concentrate on task-specific features.
\fi

\section{Conclusion}
\ifarxiv
In this work, we introduce SAE-FT, a robust fine-tuning method that leverages Sparse Autoencoders (SAEs) to regularize representation learning. The method is based on the observation that limiting the geometric drift of the vision representations of CLIP models improves fine-tuning. In contrast to standard norm regularization, SAE-FT constrains updates to an interpretable feature basis. It achieves state-of-the-art robustness on distribution shifts and superior generalization to downstream tasks compared to existing methods, especially compared to other vision encoder-only methods. Compared to a direct regularization of the geometric drift, the performance gains are marginal, but SAE-FT yields fine-grained control and makes the fine-tuning process more interpretable.

Our analysis reveals that the effectiveness of SAE-FT stems from its ability to selectively focus on task-specific features. SAE-FT re-weights pre-existing, interpretable concepts, amplifying those critical for the target task and dampens irrelevant variations. This demonstrates that interpretable regularization can enable efficient fine-tuning without erasing the learned concepts of the pre-trained model. Compared to standard norm regularization, SAE-FT gives finer control over the change and addition of features, yielding strong performance even for tasks that are difficult for the zero-shot model.

Future work could explore the regularization of both modalities, expanding the FLYP fine-tuning protocol. Applying regularization via SAEs to both the vision and text encoder could preserve features in both modalities and further improve the robustness and generalization of the fine-tuned model.
\fi

\begin{ack}
This work was supported by Institute for Information \& communications Technology Planning \& Evaluation(IITP)grant funded by the Korea government(MSIT) (RS-2019-II190075, Artificial Intelligence Graduate School Program(KAIST)).

The authors thank the International Max Planck Research School for Intelligent Systems (IMPRS-IS) for supporting Ankit Sonthalia and Arnas Uselis. 
\end{ack}

\newpage

\bibliographystyle{plainnat}
\bibliography{paper}

\appendix

\section{CKA analysis details} \label{apdx:cka}
The similarity metric CKA, proposed by \citet{kornblithSimilarityNeuralNetwork2019a}, is a similarity metric that is invariant to orthogonal projections and isotropic scaling, but not to invertible linear functions.

Assuming $X$ and $Y$ are centered it holds true that:
\begin{align}
    \frac{1}{(n-1)^2}\text{tr}(XX^TYY^T) = \|\text{cov}(X^T, Y^T)\|^2_F.
\label{eq: covariance}\end{align}
HSIC generalizes this to inner products from reproducing kernel Hilbert spaces. For two given kernels $k$ and $l$ let $K_{ij}=k(x_i, x_j)$ and $L_{ij} = l(y_i, y_j)$. The empirical estimator of HSIC is
\begin{align}
    \text{HSIC}(K, L) = \frac{1}{(n-1)^2}\text{tr}(KHLH),
\end{align}
where H is the centering matrix $H_n = I_n - \frac{1}{n}11^T$. We choose $k$ and $l$ as the linear kernels $k(x,y)=l(x,y)=x^Ty$, for which HSIC is equivalent to (\ref{eq: covariance}).

As \citet{kornblithSimilarityNeuralNetwork2019a} argue, a similarity index should be invariant to isotropic scaling, they normalize HSIC, which is known an the centered kernel alignment
\begin{align}
    \text{CKA}(K,L) = \frac{\text{HSIC}(K, L)}{\sqrt{\text{HSIC}(K, K)\text{HSIC}(L,L)}}.
\end{align}
For the experiments in section \ref{sec: drift} we use the normalized vision representations. The representations are calculated for the ImageNet test set.

\begin{table}[h]
\centering\small\setlength{\tabcolsep}{3pt}
\caption{CKA similarity matrix for vision encoder representations. Rep. avg. denotes the element-wise average of the Zero-shot and Fine-tuned activation vectors.}
\begin{tabularx}{\columnwidth}{l Y Y Y Y}
\toprule
& Zero-shot & Fine-tuned & Rep. avg. & WiSE-FT \\
\midrule
Zero-shot  & 1.00 & 0.40 & 0.67 & 0.83 \\
Fine-tuned &      & 1.00 & 0.82 & 0.59 \\
Rep. avg.  &      &      & 1.00 & 0.93 \\
WiSE-FT    &      &      &      & 1.00 \\
\bottomrule
\end{tabularx}
\label{tab: cka comparison appendix}
\end{table}

\section{FVU and feature overlap details} \label{apdx:fvu}
To compare the error the SAE makes for different representations we use Fraction of Variance Unexplained (FVU). FVU provides a normalized measure of the residual error, defined as the ratio of the Mean Squared Error (MSE) to the total variance of the dataset:
\begin{equation}
    \text{FVU} = \frac{\text{MSE}(X, \hat{X})}{\text{Var}(X)} = \frac{\frac{1}{n} \sum_{i=1}^{n} \|x_i - \hat{x}_i\|^2_2}{\frac{1}{n} \sum_{i=1}^{n} \|x_i - \bar{x}\|^2_2}
\end{equation}
where $\bar{x} = \frac{1}{n} \sum_{i=1}^{n} x_i$ is the sample mean of the original data.

While an FVU of 0 indicates a perfect reconstruction and an FVU of 1 indicates a model that performs no better than a constant prediction of the empirical mean $\bar{x}$, it is possible for the metric to exceed 1. This occurs when the MSE of the reconstruction is strictly greater than the variance of the data:
\begin{equation}
    \sum_{i=1}^{n} \|x_i - \hat{x}_i\|^2_2 > \sum_{i=1}^{n} \|x_i - \bar{x}\|^2_2.
\end{equation}

Feature overlap measures the average ratio of features that occur in both SAE features of two different representation, to the number of overall active features.

\section{Feature importance -- task relevance correlation} \label{apdx:task_relevance}
To quantitatively assess whether SAE-FT re-weights features toward task-relevant directions, we introduce a metric that measures the alignment between the active SAE features and the correct class embedding.

Each column of the SAE decoder defines a direction in the CLIP representation space. We can therefore measure how well the active features of a given sample align with the embedding of its ground-truth class. For the SAE with decoder $W \in \mathbb{R}^{d \times p}$, we denote $W_i \in \mathbb{R}^d$ as the decoder vector for the $i$-th feature. Let $s \in \mathbb{R}^{p}$ be the feature activations and $c_y \in \mathbb{R}^d$ the normalized embedding of the correct class. We define the feature-task alignment (FTA) as the activation-weighted average cosine similarity between the feature directions and the class embedding:
\begin{equation}
    \text{FTA} := \frac{\sum_{i=1}^{p} s_i \cos(W_i, c_y)}{\sum_{i=1}^{p} s_i}.
\end{equation}
FTA captures how strongly the model's active features point toward the correct class direction. A higher FTA indicates that the model places more weight on features that are aligned with the target class.

We compute the average FTA over all samples in the ImageNet test set for the zero-shot, $L_2$ regularized, and SAE-FT models. Table~\ref{tab:FTA} shows the results.

\begin{table}[h]
\centering\small\setlength{\tabcolsep}{3pt}
\caption{Average feature-task alignment (FTA) on the ImageNet test set. SAE-FT yields the highest alignment between active features and the correct class embedding.}
\begin{tabular}{l c}
\toprule
\textbf{Model} & \textbf{FTA} \\
\midrule
Zero-shot & 0.058 \\
$L_2$ regularization & 0.071 \\
\midrule
\textbf{SAE-FT} & \textbf{0.086} \\
\bottomrule
\end{tabular}
\label{tab:FTA}
\end{table}

\begin{figure}[h]
    \centering
    \includegraphics[width=0.8\textwidth]{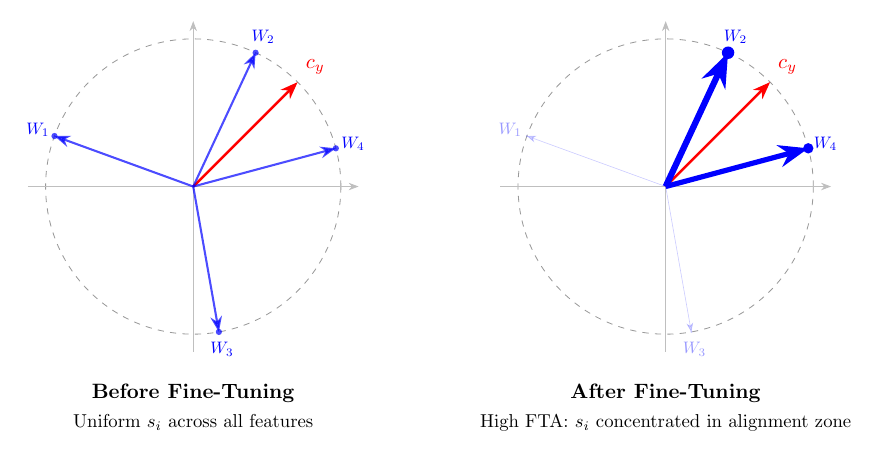}
    \caption{Visualization of the feature-task alignment metric. SAE-FT re-weights feature activations toward directions more aligned with the correct class embedding.}
    \label{fig:FTA}
\end{figure}

The results confirm that fine-tuning increases the alignment of the active features with the target class, and that SAE-FT achieves a substantially higher FTA than both the zero-shot model and $L_2$ regularization. This supports the qualitative observations in Figure~\ref{fig:feature_shift_ship}: SAE-FT adapts to the downstream task by re-weighting existing features toward directions that are more aligned with the class embeddings, rather than introducing new features. Figure~\ref{fig:FTA} provides a visualization of how the feature weighting shifts under SAE-FT.

\section{Additional feature regularization} \label{apdx:feature_reg}
We also explored regularizing the features of the SAE in a more geometrically informed way.

Let $W_d^k$ be the weight of the SAE decoder for the k-th feature. We define probability measures over the activated features of the $\text{SAE}$:
\begin{align*}
    &\nu^0 := \frac{1}{\sum_{k=1}^{d_S} s^0_k} \sum_{k=1}^{d_S} s_k^0 \delta_{W_k}\\
    &\nu^{ft} := \frac{1}{\sum_{k=1}^{d_S} s^{ft}_k} \sum_{k=1}^{d_S} s_k^{ft} \delta_{W_k}.
\end{align*}
We use optimal transport to compute the distance between different feature representations. The cost function is motivated by the original CLIP loss and uses the cosine similarity of the features
\begin{align*}
    C_{i,j} := 1 - \cos(W_d^i, W_d^j).
\end{align*}
With this we define the regularization loss as the Wasserstein distance between the two measures, given the defined cost function
\begin{align*}
    \mathcal{L}_{wass} := \mathcal{W}_1(\nu^0, \nu^{ft}; C).
\end{align*}

We used this regularization in combination with the residual regularization term, like the methods described in section \ref{sec:sae_ft}. Results are shown in Appendix \ref{apdx: additional_feature_reg_res}.

\section{PCA baseline} \label{apdx:pca}
In this section we give some additional information about the PCA baseline comparison.

As a full singular value decomposition (SVD) of the ImageNet training set is computationally expensive, we restrict ourselves to a low-rank approximation of the SVD. For our baseline we use the same number of PCA components as we have features per sample in the SAE (K=16). During fine-tuning we compute a residual and sparsity penalty similar to the SAE residual penalty.

Let $V_k \in \mathbb{R}^{d \times K}$ be the truncated right singular vector matrix. This gives the low-rank encodings $s = rV_k$ for representations $r$. With the notation of section \ref{sec:sae_ft}, we write the residual penalty as:
\begin{align}
    \mathcal{L}_{\text{resid}} := ||\Delta r - \Delta sV_k^T||_2^2.
\end{align}
We also add a similar sparsity penalty in the latent space:
\begin{align}
    \mathcal{L}_{\text{PCA}} := \lambda_{\text{resid}} \mathcal{L}_{\text{resid}} + \lambda_{\text{sparse}} ||\Delta s||_1.
\end{align}
Overall this restricts changes to the first K directions given by the SVD, with an additional penalty for changing these directions.

\section{Experiment details} \label{apdx:exp_details}

We closely follow the experimental setup of \citet{wortsmanRobustFinetuningZeroshot2022}. We fine-tune with the AdamW optimizer and a learning rate of 1e-5. The learning rate scheduler consists of a 500 steps linear warmup followed by cosine decay. Weight decay is set to 0.1 and the models are trained with a batch size of 32 for 10 epochs on one NVIDIA A100.

\subsection{SAE training}\label{apdx:sae_architecture}
For all experiments including SAE-FT, an SAE is trained on the vision representations of the zero-shot model for all images of the respective training set. We chose a Top-K SAE, which is trained for 100 epochs. The computational cost is limited, as the training of the SAE only has to be done once. The exact run-time for ImageNet is specified in Appendix~\ref{apdx:times}.

The representations the SAE is trained and evaluated on are not normalized, unless specifically mentioned. We chose the dictionary size as $4 \times d$, where $d$ is the dimension of the representation space, for all models. We chose K (the amount of active features) to be $\frac{d}{32}$, for all models.

This results in a Top-16 SAE with dictionary size 2048 for the ViT-B/16 model and a Top-24 SAE with dictionary size of 3072 for the ViT-L/14 model.

\subsection{SAE-FT training times and computational overhead} \label{apdx:times}
We report the times for the different steps needed for SAE-FT on a single NVIDIA A100 for ImageNet:
\begin{itemize}
    \item Storing representations of all training samples: 55:46 minutes
    \item Training the Top-K SAE for 100 epochs: 13:45 minutes
    \item In our testing one epoch of SAE-FT training took 1:47:23 hours, while normal fine-tuning took 1:49:34 hours. The marginal difference in time likely stems from standard variance in system I/O or background processes, suggesting that SAE-FT does not introduce computational overhead during fine-tuning of the CLIP model.
\end{itemize}
Overall when training on ImageNet for 10 epochs SAE-FT results in an increased compute time of $\sim 5\%$. This additional cost does not occur, when fine-tuning another CLIP model with the same SAE.

We additionally profile the per-step time and peak GPU memory for standard fine-tuning, $L_2$ regularization, and SAE-FT on ImageNet with ViT-B/16 on a single A100. Table~\ref{tab: compute_overhead} summarizes the results. Compared to $L_2$ regularization, the additional compute overhead of SAE-FT is negligible: per-step time increases by $0.4\%$ and peak GPU memory increases by only $19.7$\,MB due to the SAE model ($8$\,MB checkpoint). Both regularized methods require approximately $586$\,MB additional GPU memory over standard fine-tuning for the frozen encoder. Training the SAE requires storing the zero-shot representations of the training set beforehand, which amounts to $2.5$\,GB on disk for ImageNet.

\begin{table}[h]
\centering\small\setlength{\tabcolsep}{4pt}
\caption{Computational overhead comparison on ImageNet with ViT-B/16 (single NVIDIA A100). SAE-FT adds negligible cost over $L_2$ regularization.}
\begin{tabular}{l c c c}
\toprule
 & \textbf{Standard FT} & \textbf{$L_2$ Reg.} & \textbf{SAE-FT} \\
\midrule
Step time (ms) & 338.7 & 344.5 & 345.7 \\
Peak GPU memory (MB) & 6431.7 & 6997.9 & 7017.6 \\
\quad \textit{Frozen encoder} & -- & 586.2 & 586.2 \\
\quad \textit{SAE model} & -- & -- & 19.7 \\
SAE checkpoint (MB) & -- & -- & 8 \\
Precomputed repr. (GB, disk) & -- & -- & 2.5 \\
\bottomrule
\end{tabular}
\label{tab: compute_overhead}
\end{table}

\section{Further experiments}
In this section we show the performance of SAE-FT using different feature regularization methods, different hyperparameters and models.

\subsection{Hyperparameter sweep} \label{apdx:hyper_sweep}
We show the results for a hyperparameter sweep of the overall regularization parameter $\lambda$ in Table \ref{tab: hyperparameter}. With stronger regularization, the OOD performance increases, while the ID performance decreases.

\begin{table}[h]
\centering
    \caption{Hyperparameter comparison for SAE-FT. $\lambda$ is the overall regularization scale. The results for $\lambda=70$ are averaged over 3 training runs, while for all other $\lambda$ a single run is evaluated.}
    \setlength{\tabcolsep}{3pt}
    \begin{tabular}{l c c c c c c}
    \toprule
    \textbf{$\lambda$} & \multicolumn{6}{c}{\textbf{SAE-FT}} \\
    \cmidrule(lr){2-7}
    & IN & IN-R & IN-A & IN-S & IN-V2 & Avg. \\
    \midrule
    50 & \textbf{83.2} & 77.8 & 51.0 & 53.0 & \textbf{74.3} & 64.3 \\
    60 & 83.1 & 78.3 & 51.9 & \textbf{53.5} & 74.0 & 64.4 \\
    65 & 83.1 & 78.5 & 52.1 & 53.4 & 74.0 & 64.5 \\
    \textbf{70} & 82.9 & 78.5 & 52.6 & 53.4 & 73.9 & 64.6 \\
    75 & 82.8 & 78.7 & \textbf{52.9} & 53.3 & 73.9 & \textbf{64.7}\\
    90 & 82.5 & \textbf{78.9} & 52.8 & 53.4 & 73.5 & \textbf{64.7}\\
    \bottomrule
    \end{tabular}%
    \label{tab: hyperparameter}
\end{table}

\subsection{SAE architecture ablation} \label{apdx:sae_ablation}
We investigate the sensitivity of SAE-FT to the architecture of the underlying Sparse Autoencoder by varying the number of active features $K$ and the dictionary size multiplier (mult), where the dictionary size is $\text{mult} \times d$ for representation dimension $d$. Results are shown in Table~\ref{tab: sae_ablation}. Importantly, the regularization hyperparameters $\lambda_{\text{res}}$ and $\lambda_{\text{add}}$ are kept fixed at the values optimized for the default configuration ($K=16$, $\text{mult}=4$) and are not re-tuned for each SAE setup.

\begin{table}[h]
\centering
\small
\setlength{\tabcolsep}{3pt}
\caption{SAE architecture ablation for SAE-FT on ImageNet and distribution shifts (ViT-B/16). $K$ is the number of active features in the Top-K SAE, mult is the dictionary size multiplier relative to the representation dimension $d$. The default configuration is highlighted in bold.}
\begin{tabular}{c c c c c c c c}
\toprule
\textbf{$K$} & \textbf{mult} & \textbf{dict size} & \textbf{IN} & \textbf{IN-R} & \textbf{IN-A} & \textbf{IN-S} & \textbf{IN-V2} \\
\midrule
8  & 4 & 2048 & 81.8 & 75.6 & 48.6 & 51.6 & 72.9 \\
16 & 2 & 1024 & 82.4 & 76.0 & 48.4 & 51.8 & 72.5 \\
\textbf{16} & \textbf{4} & \textbf{2048} & \textbf{82.9} & \textbf{78.5} & \textbf{52.6} & \textbf{53.4} & \textbf{73.9} \\
16 & 8 & 4096 & 81.6 & 76.0 & 48.3 & 52.2 & 73.2 \\
32 & 4 & 2048 & 82.6 & 72.0 & 46.3 & 51.3 & 73.1 \\
\bottomrule
\end{tabular}
\label{tab: sae_ablation}
\end{table}

The results show that SAE-FT is sensitive to the SAE architecture, particularly on distribution shift benchmarks. In-distribution accuracy on ImageNet remains relatively stable across configurations (ranging from $81.6$ to $82.9$), but out-of-distribution performance varies substantially: IN-A ranges from $46.3$ to $52.6$ and IN-R from $72.0$ to $78.5$.

When varying the dictionary multiplier with $K=16$ fixed, both a smaller dictionary ($\text{mult}=2$) and a larger one ($\text{mult}=8$) degrade OOD performance relative to the default ($\text{mult}=4$). A dictionary that is too small likely lacks the capacity to capture the full range of semantic concepts, while an overly large dictionary may introduce redundant or poorly learned features that weaken the regularization signal. The sensitivity to $K$ follows a similar pattern: fewer active features ($K=8$) provide too coarse a representation of each sample, while more active features ($K=32$) dilute the regularization across too many directions, substantially degrading OOD robustness. We note that the performance of the configuration with $K=32$ may improve with re-tuned hyperparameters, as a larger number of active features changes the effective strength of the feature-addition penalty.

Overall, the method is most sensitive to the dictionary multiplier and the number of active features on the OOD benchmarks, while ID performance degrades only mildly. The default configuration ($K = d/32$, $\text{mult}=4$) consistently achieves the best results across all metrics.

\subsection{Additional feature regularization} \label{apdx: additional_feature_reg_res}
We also compare the other feature regularization methods discussed in Section \ref{sec:sae_ft} and Appendix \ref{apdx:feature_reg}. We show the results for simple sparse regularization (sparse), wasserstein regularization (wass) compared to standard regularization of feature addition (add) in Table \ref{tab: feature_reg}.

The restriction of feature addition outperforms other methods of SAE regularization. While wasserstein and sparsity regularization match the ID performance of addition regularization, they do not match the average accuracy on the evaluated distribution shifts.

\begin{table}[h]
\centering
    \caption{Comparisons of different SAE-FT regularization methods on ImageNet and its distribution shifts.}
    \setlength{\tabcolsep}{3pt}
    \begin{tabular}{l c c c c c c}
    \toprule
    \textbf{Method} & \multicolumn{6}{c}{\textbf{SAE-FT}} \\
    \cmidrule(lr){2-7}
    & IN & IN-R & IN-A & IN-S & IN-V2 & Avg. \\
    \midrule
    sparse & \textbf{83.0} & 77.5 & 50.7 & 52.5 & \textbf{73.8} & 63.6 \\
    wass & \textbf{83.0} & 77.9 & 51.8 & 52.8 & \textbf{74.1} & 64.2 \\
    \textbf{add} & 82.9 & \textbf{78.5} & \textbf{52.6} & \textbf{53.4} & 73.9 & \textbf{64.6} \\
    \bottomrule
    \end{tabular}%
    \label{tab: feature_reg}
\end{table}

\subsection{Further comparisons to $L_2$ regularization baseline} \label{apdx: comp_l2}
In addition to Section \ref{sec: exp_base} we provide further comparisons to generic $L_2$ regularization of the representations.

$L_2$ regularization preserves the representations of the zero-shot model and its representations generalize to other downstream datasets. Table \ref{tab: generalization_base} shows that $L_2$ regularization matches the performance of SAE-FT.

\begin{table}[h]
\centering\small\setlength{\tabcolsep}{3pt}
\caption{Generalization performance of ViT-B/16 models fine-tuned on ImageNet and evaluated on downstream transfer benchmarks.}
\begin{tabularx}{0.5\columnwidth}{l Y Y Y Y Y}
\toprule
\textbf{Method} & \multicolumn{5}{c}{\textbf{ViT-B/16}} \\
\cmidrule(lr){2-6}
& C-10 & C-100 & Cal101 & STL10 & Avg. \\
\midrule
Zero-shot & 90.8 & 68.2 & 89.6 & 98.3 & 86.7 \\
$L_2$ & \textbf{91.9} & 70.4 & \textbf{90.0} & \textbf{98.7} & \textbf{87.8} \\
\midrule
\textbf{SAE-FT} &
\makecell{\textbf{91.9} \\ {\small $\pm 0.2$}} &
\makecell{\textbf{71.2} \\ {\small $\pm 0.4$}} &
\makecell{89.5 \\ {\small $\pm 0.1$}} &
\makecell{\textbf{98.7} \\ {\small $\pm 0.0$}} &
\makecell{\textbf{87.8} \\ {\small $\pm 0.2$}} \\
\bottomrule
\end{tabularx}
\label{tab: generalization_base}
\end{table}

To assess adaptability to specialized, low-resource domains beyond the natural object categories found in ImageNet, we additionally evaluate on the Describable Textures Dataset (DTD) \cite{cimpoi2014describing}. DTD is a fine-grained texture recognition benchmark comprising 5,640 images across 47 categories. Results are shown in Table \ref{tab: dtd}.

SAE-FT and $L_2$ regularization both outperform standard fine-tuning. SAE-FT slightly outperforms the $L_2$ regularization baseline, confirming its viability for fine-tuning on domains beyond ImageNet.

\begin{table}[h]
\centering\small\setlength{\tabcolsep}{3pt}
\caption{Fine-tuning results on the Describable Textures Dataset (DTD) for ViT-B/16.}
\begin{tabular}{l c}
\toprule
\textbf{Method} & \textbf{Accuracy (\%)} \\
\midrule
Standard fine-tuning & 77.82 \\
$L_2$ regularization & 78.99 \\
\midrule
\textbf{SAE-FT} & \textbf{79.20} \\
\bottomrule
\end{tabular}
\label{tab: dtd}
\end{table}

\subsection{Results with additional models} \label{apdx:res_datasets}

SAE-FT matches the performance of state of the art methods for the larger OpenAI ViT-L/14 model on ImageNet (Table \ref{tab: ViT-L/14}). SAE-FT achieves the second highest ID accuracy behind CaRot and the second highest average accuracy on the distribution shifts behind StarFT.

\begin{table}[h]
\centering\small\setlength{\tabcolsep}{3pt}
\caption{Robust fine-tuning results on ImageNet and distribution shift benchmarks for the OpenAI ViT-L/14 model. Results for SAE-FT are averaged over 3 random seeds; subscripts denote the standard deviation.}
\begin{tabularx}{0.7\columnwidth}{l Y Y Y Y Y Y}
\toprule
\textbf{Method} & \multicolumn{6}{c}{\textbf{ViT-L/14}} \\
\cmidrule(lr){2-7}
& IN & IN-R & IN-A & IN-S & IN-V2 & Avg. \\
\midrule
Zero-shot & 75.6 & 87.9 & 70.8 & 59.6 & 69.9 & 72.0 \\
FT & 84.7 & 75.4 & 55.7 & 54.4 & 75.3 & 65.2 \\
FLYP & 86.2 & 83.8 & 68.9 & 60.2 & 78.2 & 72.8 \\
CAR-FT & 86.3 & 84.2 & 66.6 & 60.0 & 76.8 & 71.9 \\
CaRot & \textbf{87.0} & 88.0 & 72.7 & 62.7 & \textbf{79.3} & 75.6 \\
WiSE-FT & 86.1 & 88.5 & 72.9 & \textbf{63.6} & 78.1 & 75.8 \\
StarFT & 86.4 & 88.7 & \textbf{73.8} & 63.2 & 78.9 & \textbf{76.2} \\
\midrule
\textbf{SAE-FT} &
\makecell{86.5 \\ {\small $\pm 0.1$}} &
\makecell{\textbf{88.8} \\ {\small $\pm 0.0$}} &
\makecell{73.1 \\ {\small $\pm 0.1$}} &
\makecell{63.5 \\ {\small $\pm 0.1$}} &
\makecell{78.6 \\ {\small $\pm 0.1$}} &
\makecell{76.0 \\ {\small $\pm 0.0$}} \\
\bottomrule
\end{tabularx}
\label{tab: ViT-L/14}
\end{table}

We evaluate SAE-FT on the ViT-B/16 SigLIP2 \cite{tschannen2025siglip2multilingualvisionlanguage} model. The results are shown in Table \ref{tab: sig_lip}, comparing SAE-FT to WiSE-FT. For WiSE-FT the optimal $\alpha$ (0.5) is chosen and for SAE-FT we do a hyperparameter search for the scale of the overall regularization (optimal $\lambda=70$). SAE-FT outperforms WiSE-FT, by achieving a higher ID accuracy by $0.5$ percentage points and a higher average accuracy across the distribution shifts.

\begin{table}[h]
\centering
\caption{Results for the SigLIP2 ViT-B/16 model on ImageNet.}
\begin{tabular}{l c c c c c c}
\toprule
\textbf{Method} & \multicolumn{6}{c}{\textbf{SigLIP2 ViT-B/16}} \\
\cmidrule(lr){2-7}
& IN & IN-R & IN-A & IN-S & IN-V2 & Avg. \\
\midrule
Zero-shot & 78.5 & \textbf{91.7} & 55.2 & \textbf{68.9} & 71.3 & 71.8 \\
FT & 82.8 & 75.1 & 37.4 & 57.2 & 72.3 & 60.5 \\
WiSE-FT & 84.8 & 89.5 & \textbf{56.2} & \textbf{68.9} & 76.5 & 72.8  \\
\midrule
\textbf{SAE Add} & \textbf{85.3} & 90.2 & \textbf{56.2} & 68.6 & \textbf{77.1} & \textbf{73.0} \\
\bottomrule
\end{tabular}
\label{tab: sig_lip}
\end{table}

\newpage

\section{Further qualitative results} \label{apdx:res_qual}
In this section we show further qualitative results of samples from the ImageNet test analyzed with the SAE used in the SAE-FT regularization. We focus on samples for which the predictions and features of the zero-shot and fine-tuned model differ. We also include (rare) examples of samples, which SAE-FT misclassified that are correctly classified by the zero-shot model.

\begin{center}
\includegraphics[width=0.8\textwidth]{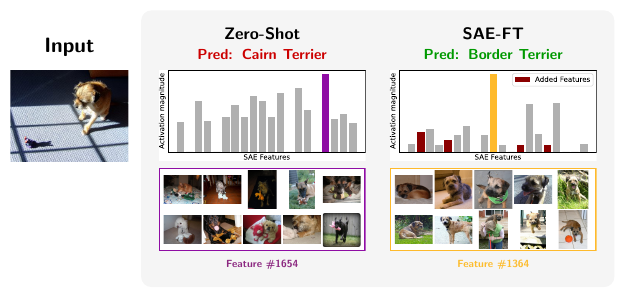}
\par\vspace{1em}
\includegraphics[width=0.8\textwidth]{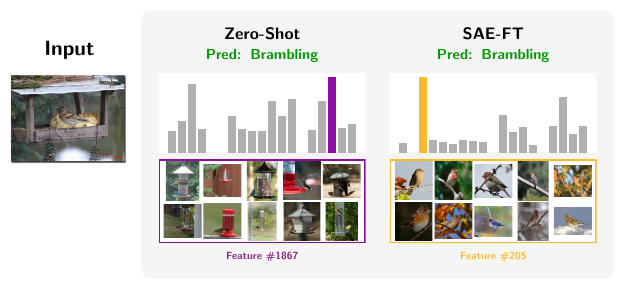}
\par\vspace{1em}
\includegraphics[width=0.8\textwidth]{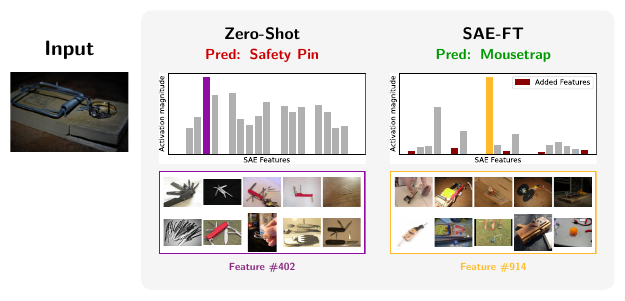}
\par\vspace{1em}
\includegraphics[width=0.8\textwidth]{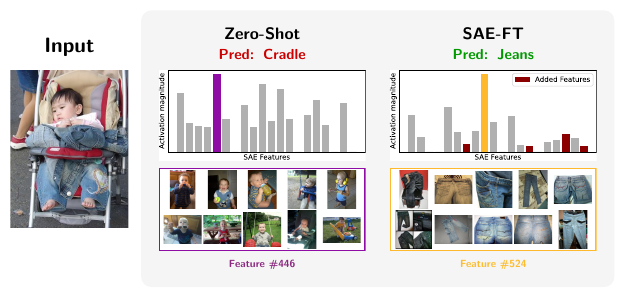}
\par\vspace{1em}
\includegraphics[width=0.8\textwidth]{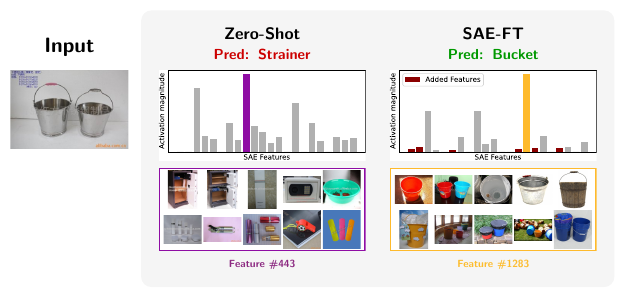}
\par\vspace{1em}
\includegraphics[width=0.8\textwidth]{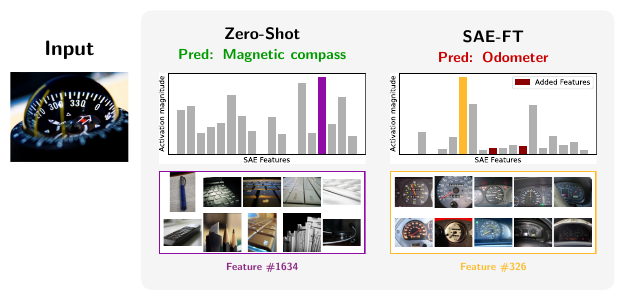}
\par\vspace{1em}
\includegraphics[width=0.8\textwidth]{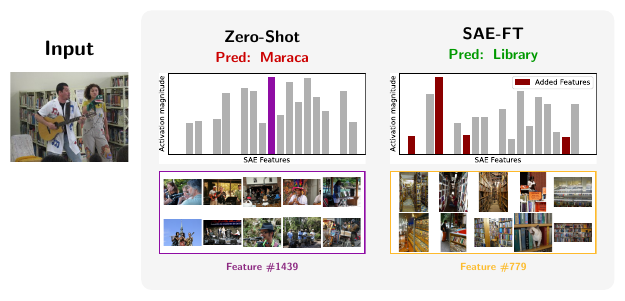}
\par\vspace{1em}
\includegraphics[width=0.8\textwidth]{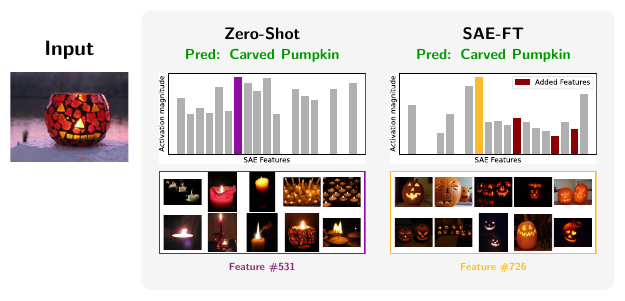}
\par\vspace{1em}
\includegraphics[width=0.8\textwidth]{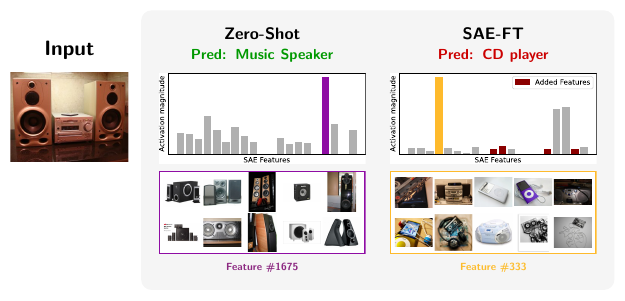}
\par\vspace{1em}
\includegraphics[width=0.8\textwidth]{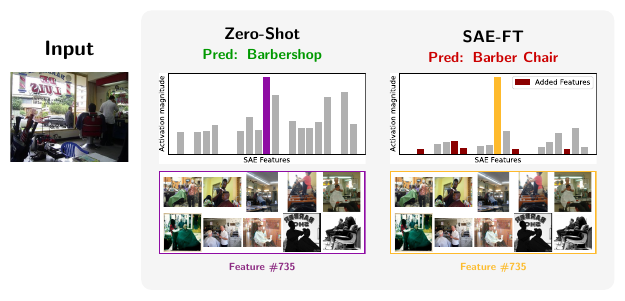}
\end{center}

\end{document}